\documentclass[lettersize,journal]{IEEEtran}
\usepackage[table,xcdraw]{xcolor}
\usepackage{amsmath,amsfonts}
\usepackage{algorithmic}
\usepackage{algorithm}
\usepackage{array}
\usepackage[caption=false,font=normalsize,labelfont=sf,textfont=sf]{subfig}
\usepackage{textcomp}
\usepackage{stfloats}
\usepackage{url}
\usepackage{verbatim}
\usepackage{graphicx}
\usepackage{cite}
\usepackage{orcidlink}

\usepackage{booktabs}
\usepackage{geometry}
\usepackage{wrapfig} 

\usepackage[export]{adjustbox} 

\begin{document}

\title{Interpretable Few-Shot Image Classification via Prototypical Concept-Guided Mixture of LoRA Experts}

\author{Zhong Ji \orcidlink{0000-0002-2197-3739},~\IEEEmembership{Senior Member,~IEEE,}
        Rongshuai Wei \orcidlink{0009-0007-8117-5361}, 
        Jingren Liu \orcidlink{0009-0009-0163-4105},
        Yanwei Pang \orcidlink{0000-0001-6670-3727},~\textit{Senior Member,~IEEE,}
        Jungong Han \orcidlink{0000-0003-4361-956X},~\textit{Senior Member,~IEEE} 
\thanks{This work was supported by the National Natural Science Foundation of China (NSFC) under Grants 62176178 and 62441235. (Corresponding author: Jingren Liu.)}
\thanks{Zhong~Ji and Yanwei~Pang are with the School of Electrical and Information Engineering, Tianjin Key Laboratory of Brain-Inspired Intelligence Technology, Tianjin University, Tianjin 300072, China, and also with the Shanghai Artificial Intelligence Laboratory, Shanghai 200232, China (e-mail: \{jizhong, pyw\}@tju.edu.cn).}
\thanks{Rongshuai Wei and Jingren Liu are with the School of Electrical and Information Engineering, Tianjin Key Laboratory of Brain-Inspired Intelligence Technology, Tianjin University, Tianjin 300072, China (e-mail: 2023234097@tju.edu.cn; jrl0219@tju.edu.cn).}
\thanks{Jungong Han is with the Department of Automation, Tsinghua University, Beijing 100084, China (e-mail: jungonghan77@gmail.com).}}

\markboth{Journal of \LaTeX\ Class Files,~Vol.~14, No.~8, August~2021}%
{Shell \MakeLowercase{\textit{et al.}}: A Sample Article Using IEEEtran.cls for IEEE Journals}


\maketitle

\begin{abstract} 
Self-Explainable Models (SEMs) rely on Prototypical Concept Learning (PCL) to enable their visual recognition processes more interpretable, but they often struggle in data-scarce settings where insufficient training samples lead to suboptimal performance. To address this limitation, we propose a Few-Shot Prototypical Concept Classification (FSPCC) framework that systematically mitigates two key challenges under low-data regimes: parametric imbalance and representation misalignment. Specifically, our approach leverages a Mixture of LoRA Experts (MoLE) for parameter-efficient adaptation, ensuring a balanced allocation of trainable parameters between the backbone and the PCL module. Meanwhile, cross-module concept guidance enforces tight alignment between the backbone’s feature representations and the prototypical concept activation patterns. In addition, we incorporate a multi-level feature preservation strategy that fuses spatial and semantic cues across various layers, thereby enriching the learned representations and mitigating the challenges posed by limited data availability. Finally, to enhance interpretability and minimize concept overlap, we introduce a geometry-aware concept discrimination loss that enforces orthogonality among concepts, encouraging more disentangled and transparent decision boundaries. Experimental results on six popular benchmarks (CUB-200-2011, \emph{mini}-ImageNet, CIFAR-FS, Stanford Cars, FGVC-Aircraft, and DTD) demonstrate that our approach consistently outperforms existing SEMs by a notable margin, with 4.2\%–8.7\% relative gains in 5-way 5-shot classification. These findings highlight the efficacy of coupling concept learning with few-shot adaptation to achieve both higher accuracy and clearer model interpretability, paving the way for more transparent visual recognition systems.
\end{abstract}


\section{Introduction}
Recent advancements in Self-Explainable Models (SEMs), such as ProtoPNet~\cite{chen2019looks} and TCPL~\cite{gao2024learning}, have underscored a 
critical role of interpretability in model design~\cite{10106785,8963848,yu2024explainability}, aligning with the broader objectives of eXplainable Artificial Intelligence (XAI)~\cite{arrieta2020explainable}. These models leverage Prototypical Concept Learning (PCL) to enhance transparency while maintaining competitive performance on standard classification benchmarks. However, their ability to generalize under data-constrained scenarios remains insufficiently investigated. This limitation is particularly pronounced in Few-Shot Classification (FSC), a setting that inherently assesses a model's ability to extrapolate learned representations with minimal supervision, an essential criterion to ensure the reliability and robustness of XAI-driven explanations. To address this gap, we propose a novel Few-Shot Prototypical Concept Classification (FSPCC) framework as shown in Fig.~\ref{fig:ppc_flowchart}, systematically evaluating its generalization ability and interpretability to advance the frontiers of SEMs.

\begin{figure}[t]
    \begin{flushright}
        \centering
        \includegraphics[scale=0.9]{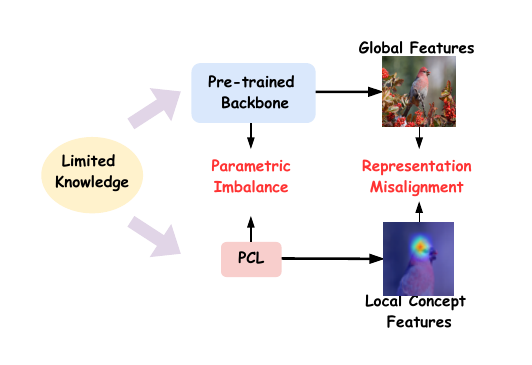}
        \caption{The key challenges limiting model generalization in FSPCC.}
        \label{fig:motivation}
    \end{flushright}
\end{figure}

Building on the existing architectures of SEMs and our empirical analysis of FSPCC, we identify key challenges that constrain model generalization, as illustrated in Fig.~\ref{fig:motivation}. A predominant issue arises from the widespread reliance on fully fine-tuning overparameterized pre-trained backbones, which induces catastrophic representational drift within foundational feature spaces while imposing substantial computational overhead. This parametric imbalance introduces an optimization bias, leading to gradient updates that disproportionately adjust backbone parameters during joint training. As a result, distortions emerge within the learned concept feature manifolds, characterized by insufficient inter-concept separation and inadequate intra-concept compactness. These deficiencies, in turn, generate semantically ambiguous cluster formations that fundamentally compromise human-aligned interpretability. This phenomenon becomes particularly pronounced in low-data regimes, where sparse supervisory signals fail to effectively regularize the high-capacity backbone, exacerbating feature space entanglement through unconstrained parametric drift. 

Furthermore, we identify a critical misalignment between pre-trained feature representations and the requirements of PCL. Standard pre-training objectives, designed to optimize holistic classification performance, often prove inadequate for capturing fine-grained, part-specific discriminators, such as distinguishing ``beak'' from ``back'' in ornithological concepts. By prioritizing global discriminative patterns over localized concept-specific features, 
it creates an inherent incompatibility between pre-trained representations and PCL demands. Naïve fine-tuning attempts to resolve this mismatch but instead exacerbates destructive interference in concept embedding spaces, leading to prototype collapse within overlapping feature manifolds. Addressing this challenge necessitates a paradigm shift towards selective feature adaptation mechanisms that preserve the structural integrity of pre-trained knowledge while enabling dynamic specialization for part-aware concept discrimination.

\begin{figure*}[t]
    \centering  
    \includegraphics[scale=0.8]{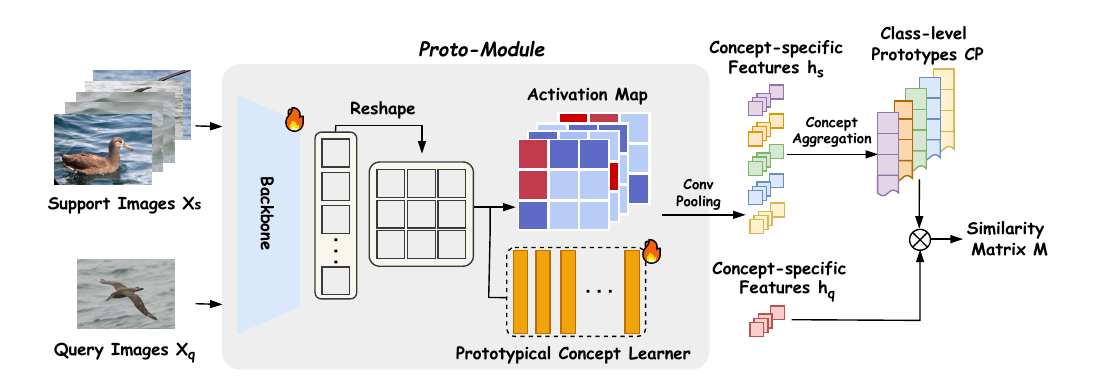}
    \caption{A comprehensive flowchart of the Few-Shot Prototypical Concept Classification framework with a prototypical concept learner.}
    \label{fig:ppc_flowchart}
\end{figure*}

\begin{figure*}[t]
    \centering  
    \includegraphics[scale=0.8]{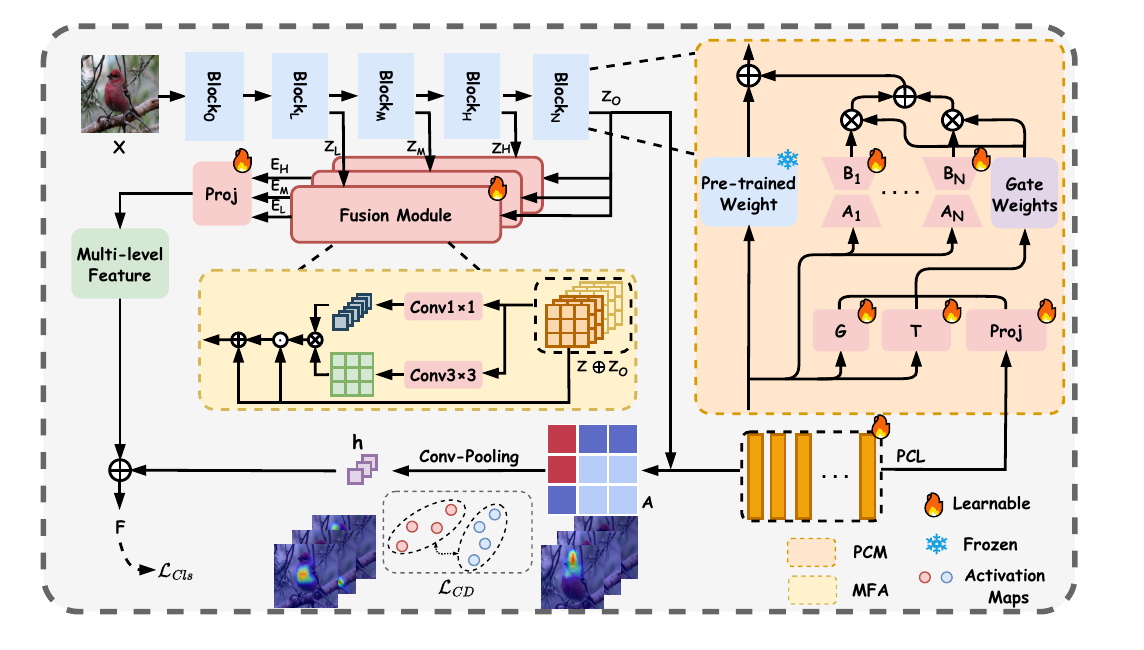} 
    \caption{Schematic diagram of the specific module composition of our FSPCC framework,including PCM module, MFA module and \(\mathcal{L}_{CD}\).}
    \label{fig:ppc_flowchart2}
\end{figure*}

To systematically tackle the two challenges 
in existing SEMs under few-shot conditions, we introduce our method named \(\mathbf{L}\)ow-rank \(\mathbf{E}\)xpert \(\mathbf{P}\)rototypical \(\mathbf{P}\)art \(\mathbf{N}\)etwork (LE-ProtoPNet) including four cohesive architectural innovations that holistically redesign model optimization pathways as shown in Fig.~\ref{fig:ppc_flowchart2}. 

First, we eschew conventional full-parameter adaptation of pre-trained backbones through a parameter-efficient tuning strategy. Our proposed Mixture of LoRA Experts (MoLE) architecture implements multiple low-rank adaptation matrices as specialized concept learners, achieving parametric equilibrium between the fixed backbone and trainable PCL module via adaptive resource allocation. This preserves self-supervised pre-trained knowledge while preventing over-parameterization through dimensionally-aligned projections.

Second, to address representation misalignment between backbone and PCL optimization spaces, we establish a direct cross-module interaction beyond implicit gradient-based connections. The PCL is structurally embedded into MoLE's gating module through dimension-preserving concatenation, enabling dynamic expert specialization aligned with concept activation patterns. This explicit fusion mechanism drives hierarchical feature refinement from class-level semantics to localized discriminative concepts through attention-weighted expert routing. 

Third, for mitigating knowledge scarcity in few-shot regimes, we develop a multi-level feature preservation framework that progressively aggregates multi-granular backbone representations, counteracting information degradation in deep processing while enhancing concept diversity through complementary feature integration. 

Finally, we formulate a concept discrimination loss that imposes geometric constraints on concept cluster distributions. By enforcing inter-concept separation, the concept discrimination loss induces structured concept manifolds that enhance both feature discriminability and model interpretability while maintaining compatibility with pre-trained knowledge priors.
In summary, our research makes two contributions:
\begin{itemize}
    \item We propose a FSPCC framework that integrates SEMs with few-shot learning to capture human-interpretable concepts under minimal supervision, ensuring both robust classification and model transparency in real-world applications.
    \item To address parametric imbalance and representation misalignment in FSPCC, we introduce our LE-ProtoPNet including four cohesive enhancements: (1) a PEFT scheme leveraging MoLE to curb backbone preference under data scarcity and effectively adaptively learn different concept sets, (2) a cross-module synergy guided by PCL module for refined feature interaction in MoLE, (3) a multi-level feature preservation mechanism that retains essential deep and shallow representations, and (4) a 
    concept discrimination loss enforcing orthogonal separability among concepts.
\end{itemize}

Extensive experimental results on few-shot benchmarks demonstrate superior classification accuracy, improved inter-concept separability, and enhanced interpretability, attesting to the effectiveness of our design choices and laying a solid foundation for further research in SEMs. 
The remainder of this paper is organized as follows. In Section~\ref{sec2}, we provide a review of the related work in the field, highlighting key approaches and their limitations. In Section~\ref{sec3}, we describe the methodology in detail, including the model framework and the practical realization. Section~\ref{sec4} outlines the experimental setup and results. Finally, Section~\ref{sec5} concludes the paper.

\section{Related Works}
\label{sec2}
\noindent\textbf{Self-Explainable Models (SEMs).} In XAI, SEMs have garnered significant attention for providing human-centric explanations\cite{liu2023prediction,li2023joint}. A representative example is ProtoPNet~\cite{chen2019looks}, which learns human-interpretable visual concepts from images and uses these concepts for classification, effectively circumventing black-box mechanisms. Building on this foundation, TesNet~\cite{wang2021interpretable} maps image parts onto a Grassmann manifold \cite{bendokat2024grassmann} to capture high-level features while enforcing an orthogonality constraint on prototype distributions, enhancing both diversity and interpretability. PIP-Net~\cite{nauta2023pip} refines this approach by incorporating human perception-aligned regularization, reducing the semantic gap between machine-learned and human-interpretable concepts. More recently, ProtoS-ViT~\cite{turbe2024protos} leverages pre-trained Vision Transformers (ViT)\cite{dosovitskiy2020image}, achieving a balance between efficiency and explainability. Despite these advancements, most SEMs have been evaluated on standard classification benchmarks, with limited exploration in more challenging zero-shot or few-shot scenarios. Such data-scarce settings are crucial for assessing model generalization beyond conventional datasets. To address this gap, we propose a dedicated few-shot SEM framework that demonstrates how interpretability-driven models adapt with minimal training data, thereby advancing the understanding of SEMs in resource-limited contexts and contributing to the evolution of the XAI field.

\noindent\textbf{Few-Shot Learning (FSL).} FSL addresses the challenge of enabling models to generalize effectively to novel tasks with minimal labeled samples, emphasizing both computational efficiency and robust generalization. Over the years, this field sees significant advancements. 
Recently, the advent of large-scale pre-trained models
revolutionizes FSL by enabling the transfer of knowledge from extensive datasets. In particular, prompt learning methods such as CoOp \cite{zhou2022learning} and its extensions \cite{zhou2022conditional,yao2023visual,zhu2023prompt} effectively adapt vision-language models (VLMs) to limited labeled data. Adapter-based methods also gain traction, with CLIP-Adapter \cite{gao2024clip} introducing visual adapters to merge adapted and original features, Tip-Adapter \cite{zhang2021tip} leveraging a query-key cache mechanism without additional training, and TaskRes \cite{yu2023task} optimizing adaptation to target tasks via prior-independent parameters. Despite these advancements, interpretability remains an underexplored aspect of FSL. To address this gap, we integrate the PCL module from SEMs into pre-trained frameworks and propose a novel Mixture of LoRA Experts (MoLE) mechanism. This approach facilitates precise learning for each concept set, providing a robust foundation for the development of transparent, interpretable, and adaptable FSL models.

\noindent\textbf{Mixture of Experts (MoE).}  
By allowing different experts to specialize in specific knowledge areas, MoE~\cite{jacobs1991adaptive} demonstrates its ability to enhance representational capacity. Early approaches~\cite{jacobs1991adaptive,jordan1994hierarchical,shahbaba2009nonlinear,collobert2001parallel} utilize dense mixture layers, where all experts are activated during each iteration, leading to significant computational overhead. To address this issue, Shazeer \textit{et al.}~\cite{shazeer2017outrageously} introduce the sparsely-gated MoE layer, which activates only a subset of experts during each forward pass. This design employs a gating network and a top-\(k\) routing strategy, which are widely adopted in subsequent studies. To ensure balanced expert utilization, GShard~\cite{lepikhin2020gshard} defines a differentiable auxiliary loss \(\mathcal{L}_{aux}\), which Switch Transformers~\cite{fedus2021switch} incorporate to simplify the MoE routing algorithm while reducing communication and computational costs. Li \textit{et al.}~\cite{li2023adaptive} propose an adaptive gating mechanism for MoE-based language models, dynamically adjusting the number of active experts based on token complexity to optimize computational efficiency. 
DeepSeekMoE~\cite{dai2024deepseekmoe} decomposes experts into smaller units, enabling flexible combinations of activated experts while maintaining computational efficiency. Distinguishing itself from these studies, our MoLE model combines MoE with Parameter-Efficient Fine-Tuning (PEFT), integrating the strengths of both frameworks to achieve superior performance while remaining computationally efficient. This approach addresses the trade-off between scalability and adaptability, offering an effective solution for resource-constrained scenarios.


\section{Method}
\label{sec3}
\subsection{\textbf{Few-Shot Prototypical Concept Classification}} 
\label{sec:FSPCC}
First, in this subsection, we detail our proposed Few-Shot Prototypical Concept Classification (FSPCC) framework, which extends the traditional Few-Shot Classification (FSC) by incorporating concept-specific feature learning to improve interpretability and classification performance. Unlike standard FSC methods that utilize holistic embeddings, FSPCC decomposes images into localized representations, allowing the model to focus on discriminative visual components. This refinement better aligns with human perception and meets the growing demand for model transparency.

The FSPCC framework partitions the dataset into base classes \( \mathcal{D}_{\text{base}} \) and novel classes \( \mathcal{D}_{\text{novel}} \), ensuring \( \mathcal{D}_{\text{base}} \cap \mathcal{D}_{\text{novel}} = \emptyset \). Training is performed on \( \mathcal{D}_{\text{base}} \) with ample labeled samples, while testing is conducted on \( \mathcal{D}_{\text{novel}} \), which contains limited annotations. Few-shot tasks are simulated as \( N \)-way \( K \)-shot settings, involving a support set \( \mathbf{X}_{\text{support}} \) and a query set \( \mathbf{X}_{\text{query}} \). A backbone network, such as a CNN (without pooling) or a ViT (without positional encoding), extracts feature embeddings \( \mathbf{F}_S \in \mathbb{R}^{N \times K \times R \times D} \) and \( \mathbf{F}_Q \in \mathbb{R}^{N \times 1 \times R \times D} \), where \( R \) represents the number of local features and \( D \) denotes the feature dimensionality. For simplicity, subscripts are omitted as the Proto-Module processing in Fig.~\ref{fig:ppc_flowchart} is identical for all inputs.

To incorporate interpretability, FSPCC employs a Prototypical Concept Learner (PCL) module that identifies visual concepts via an optimizable concept matrix \( \mathbf{P} \in \mathbb{R}^{C \times D} \), where \( C \) denotes the concept dimension. The relationship between part-specific features \( \mathbf{F}\) and concepts \( \mathbf{P} \) is measured by an activation score:
\begin{equation}
    \mathbf{A} = \frac{\langle \mathbf{F}, \mathbf{P} \rangle}{\| \mathbf{F} \|_2^2 \| \mathbf{P} \|_2^2} \in \mathbb{R}^{N (\times K) \times R \times C}.
\end{equation}

To enhance smoothness and suppress noise, the activation map \( A \) is scaled using a softmax function with a temperature parameter \( \tau \):
\begin{equation}
    \tilde{\mathbf{A}} = \mathrm{softmax} \left( \frac{\mathbf{A}}{\tau} \right)\in \mathbb{R}^{N (\times K) \times R \times C}.
\end{equation}

Next, in order to capture spatial relationships between image parts while maintaining concept independence to generate concept features $\mathbf{h} \in \mathbb{R}^{N (\times K) \times C}$, the refined activation map \( \tilde{A} \) undergoes depthwise convolutions with independent kernels:
\begin{equation}
\label{h_eq}
\small
    \mathbf{h} = \mathrm{\text{maxpool}}\{\mathrm{\text{LN}} \left( \text{Conv}_{1 \times 1}(\tilde{\mathbf{A}}) + \text{Conv}_{3 \times 3}(\tilde{\mathbf{A}}) \right)\},
\end{equation}
where $\mathrm{\text{LN}}$ refers to Layer Normalization, and $\mathrm{\text{maxpool}}$ represents the max pooling operation.

To further align with human preferences in FSC, $\mathbf{h}_s$ from the support set are aggregated into class-level prototypical concepts. For all given classes, the prototypical concept \( \mathbf{CP} \in \mathbb{R}^{N \times C} \) is computed as:
\begin{equation}
    \mathbf{CP} = \frac{1}{K} \sum_{i=1}^{K} \mathbf{h}_s^i,
\end{equation}
Here, \( \mathbf{h}_s^i \in \mathbb{R}^{N \times 1 \times C} \) denotes the concept-specific feature of the \( i \)-th support sample, with these prototypical concepts serving as reference points for query samples.

Specifically, the concept-specific features of query samples \( \mathbf{h}_q \in \mathbb{R}^{N \times C} \) are compared to the class-level prototypes \( \mathbf{CP} \) using cosine similarity, resulting in a similarity matrix \( \mathbf{M} \in N \times N \):
\begin{equation}
    \mathbf{M} = \frac{\mathbf{h}_q \cdot \mathbf{CP}}{\| \mathbf{h}_q \|_2 \| \mathbf{CP} \|_2}.
\end{equation}

The predicted pseudo-labels for query samples in the few-shot setting are determined by selecting the class corresponding to the highest similarity scores, formulated as \(\mathbf{Y}_q = \arg\max(\mathbf{M}) \in \mathbb{R}^{N \times 1}\). This subsection introduces the integration of Prototypical Concept Classification into the few-shot framework, drawing significant inspiration from the principles of SEMs~\cite{chen2019looks, wang2021interpretable, nauta2023pip, turbe2024protos}. However, as SEMs are not inherently designed to address FSL, their direct application remains suboptimal. To overcome this limitation, we propose a reformulated framework with four primary objectives: (1) leveraging the pre-trained model and the MoLE structure to mitigate parameter imbalance, enabling each expert to adaptively learn a subset of concepts, (2) refining model parameters through concepts to minimize feature mismatches and enhance interpretability, (3) maximizing the extraction of knowledge from limited samples to mitigate the disadvantages of FSL, and (4) enforcing strict separability between concepts to enhance human-aligned interpretability. This reformulation bridges the gap between interpretability and generalization, enabling a more effective deployment of SEM-inspired methodologies within the constraints of FSL. 


\subsection{\textbf{Incorporating PCL Module into MoLE}}
While recent advancements in SEMs~\cite{nauta2023pip, turbe2024protos} have begun leveraging pre-trained models to enhance performance, we contend that their potential remains insufficiently realized. Specifically, current SEMs~\cite{wang2021interpretable, nauta2023pip, turbe2024protos} primarily rely on a full fine-tuning paradigm, even when utilizing state-of-the-art self-supervised pre-trained weights (DINO~\cite{caron2021emerging} and DINO-V2~\cite{oquab2023dinov2}). However, recent developments in PEFT \cite{hu2021lora,zhang2023adalora,lester2021power,li2021prefix,ding2023sparse,dong2024enhancing} have revealed that selectively tuning a subset of parameters can, in certain cases, outperform traditional full fine-tuning, maximize the generalization capabilities of pre-trained models, significantly mitigate parameter imbalance in FSPCC, and reduce computational overhead. Building on these insights, we introduce PEFT into the SEMs for the first time, seamlessly integrating it with PCL optimization. This integration forms the foundation of our Prototypical Concept-guided Mixture of LoRA Experts components, which are designed to achieve superior efficiency and enhanced generalization, particularly in few-shot scenarios.

\subsubsection{\textbf{Mixture of LoRA Experts (MoLE)}} \label{sec:MoLE_impl}
First, we describe the pipeline of our basic MoLE component. As illustrated in Fig.~\ref{fig:ppc_flowchart2}, MoLE avoids fine-tuning all parameters in a large backbone network by introducing multiple low-rank adaptation (LoRA) \cite{hu2021lora} modules, each termed an expert. For a given linear projection or attention weight matrix \(\mathbf{W}_{0}\), each LoRA expert inserts two trainable low-rank matrices \(\mathbf{A}_{i}, \mathbf{B}_{i}\) of rank \(r\), generating a rank-constrained update \(\Delta \mathbf{W}_{i} = \mathbf{B}_{i}\mathbf{A}_{i}\). The ensemble of experts, \(\{\Delta \mathbf{W}_{i}\}_{i=1}^{E}\) (\(E\) denotes the number of experts), captures diverse adaptation paths while keeping the original pre-trained knowledge. The summation of the base weight plus a chosen expert’s update forms the effective layer-specific weight:
\begin{equation}
    \mathbf{W} = \mathbf{W}_{0} + \alpha \Delta \mathbf{W}_{i},
\end{equation}
where \(\alpha\) scales the magnitude of the update and \(i \in \{1,\ldots,E\}\) indexes the active expert.

Although this ensemble-based method is effective, it can easily lead to homogeneous learned features. Therefore, to enable intelligent and adaptive selection of different experts for the inputs, we propose a gating mechanism, which dynamically determines the activation of LoRA experts. Specifically, the gating network adaptively makes decisions based on the intermediate variable \(\mathbf{z} \in \mathbb{R}^{N \times R \times D}\) within each backbone block and projects \(\mathbf{z}\) into a lower-dimensional space through a Multi-Layer Perceptron (MLP), denoted as \(\mathbf{G}\):
\begin{equation}
    \mathbf{g} = \text{softmax}(\mathbf{G}(\mathbf{z})) \in \mathbb{R}^{N \times R \times E},
\end{equation}
where \(R\) represents the number of parts or patches in the intermediate layer of the images.


During training, the top-\(k\) experts, corresponding to the highest-ranked indices in \(\mathbf{g}\), are typically selected from the \(E\) available experts. However, the relative importance of these selected experts remains ambiguous. To address this, we introduce a contribution function to quantify the importance score of each expert, \(\mathbf{e}\), in the output. To further reduce interference, we incorporate an additional MLP, denoted as \(\mathbf{T}\), for refinement.
\begin{equation}
\label{filter1}
    \mathbf{\epsilon} = \frac{1}{E} \times \mathrm{sigmoid}(\mathbf{T}(\mathbf{z})) \in \mathbb{R}^{N \times R \times 1},
\end{equation}
\begin{equation}
\label{filter2}
    \mathbf{e} = 
    \begin{cases}
    \mathbf{g} - \mathbf{\epsilon} & \text{if } \mathbf{g} \ge \mathbf{\epsilon}, \\
    0 & \text{otherwise},
    \end{cases}
\end{equation}
where \(\epsilon\) serves as a trainable cutoff threshold. 

Once the importance scores \(\mathbf{e}\) are computed, the MoLE outputs \(\mathbf{z}_{\mathrm{MoLE}}\), which encapsulates richer intermediate-layer information and multi-faceted knowledge, can be obtained by combining each expert’s low-rank transformation of \(\mathbf{z}\):
\begin{equation}
\small
   \mathbf{z}_{\mathrm{MoLE}}
   \;=\;
   \sum_{i=1}^{E}
   \underbrace{\frac{\,\mathbf{e}_i \,}{\sum_{j=1}^E \bigl(\mathbf{e}_j\bigr)}}_{\text{Gating-and-Importance}}
   \;\odot\;
   \Bigl(\mathbf{B}_{i}\,\mathbf{A}_{i}\,\mathbf{z}\Bigr),
\label{mole}
\end{equation}
where \(\odot\) denotes broadcast (elementwise) multiplication along the appropriate dimensions, and \(\sum_{j=1}^E \bigl(\mathbf{e}_j \odot \mathbf{g}_j\bigr)\) ensures a proper  normalization over experts. 

By strategically selecting and weighting relevant low-rank experts, MoLE effectively adapts a pre-trained model to few-shot tasks with minimal trainable parameters. However, the connection between this process and PCL remains limited. To address this, we refine the pipeline to enable each concept in PCL to be adaptively and intelligently learned by different LoRA experts, culminating in the Prototypical Concept-guided MoLE framework described below.

\subsubsection{\textbf{Prototypical Concept-guided MoLE (PCM) }} 
\label{sec:PCg_MoLE_impl}
Building on the MoLE foundation, we integrate it with PCL from Section~\ref{sec:FSPCC}, forming the Prototypical Concept-guided MoLE. While PCL aligns part-specific or patch-level features with learned concepts \(\mathbf{P}\), its current implementation is coarse. Directly using raw image parts and \(\mathbf{P}\) to generate self-explainable features, followed by loss computation and backpropagation, only indirectly optimizes the MoLE module, preventing individual experts from learning specific concepts. To address this, we refine Eq.~\ref{mole} to allow PCL to directly guide the learning of distinct experts.

Recall that the original MoLE gating vector \(\mathbf{g}\in\mathbb{R}^{N\times R\times E}\) (Sec.~\ref{sec:MoLE_impl}) is derived from the the intermediate features \(\mathbf{z}\). To achieve MoLE learning constrained by concepts, we propose an attention-based concept alignment step that refines \(\mathbf{g}\) via the concepts \(\mathbf{P}\). Specifically, we compute an attention score between \(\mathbf{g}\) and \(\mathbf{P}\), which yields:
\begin{equation}
    \mathbf{g}' 
    \;=\;
    \mathrm{softmax}\!\Bigl(\mathrm{Attn}\!\bigl(\mathbf{g},\mathbf{P}\bigr)\Bigr),
    \quad
    \mathbf{g}'\in\mathbb{R}^{N\times R\times E},
\label{eq:g_concept}
\end{equation}
where \(\mathrm{Attn}(\cdot)\) is a learnable operator (e.g., a scaled dot-product attention) that projects \(\mathbf{g}\) onto the concept space. In essence, \(\mathbf{g}'\) captures how much each expert aligns with each of the prototypical concepts.

Then, we fuse the original MoLE gating \(\mathbf{g}\) with the concept-driven attention \(\mathbf{g}'\) to obtain an enriched gate:
\begin{equation}
    \tilde{\mathbf{g}}
    \;=\;
    \mathbf{g} + \mathbf{g}',
    \quad
    \tilde{\mathbf{g}}\in\mathbb{R}^{N\times R\times E}.
\label{eq:g_fused}
\end{equation}

Subsequently, by applying Eq.~\ref{filter1} and Eq.~\ref{filter2} to this new activation $\tilde{\mathbf{g}}$, we derive the updated Prototypical Concept-guided MoLE features as follows:
\begin{equation}
\label{filter3}
    \tilde{\epsilon}
    = \frac{1}{E} \times \mathrm{sigmoid}\bigl(\mathbf{T}(\mathbf{z})\bigr)
    \;\in\; \mathbb{R}^{N \times R \times 1},
\end{equation}

\begin{equation}
\label{filter4}
    \tilde{\mathbf{e}} = 
    \begin{cases}
        \tilde{\mathbf{g}} - \tilde{\epsilon}, & \text{if } \tilde{\mathbf{g}} \ge \tilde{\epsilon}, \\
        0, & \text{otherwise}.
    \end{cases}
\end{equation}

\begin{equation}
\small
   \tilde{\mathbf{z}}_{\mathrm{MoLE}}
   \;=\;
   \sum_{i=1}^{E}
   \underbrace{ \frac{\tilde{\mathbf{e}}_i}{\sum_{j=1}^E \tilde{\mathbf{e}}_j} }_{\substack{\text{Prototypical Concept-guided}\\ \text{Gating-and-Importance}}}
   \;\odot\;
   \Bigl(\mathbf{B}_{i}\,\mathbf{A}_{i}\,\mathbf{z}\Bigr),
\label{eq:z_mole_proto}
\end{equation}




By explicitly integrating the MoLE with concept alignment, the Prototypical Concept-guided MoLE dynamically routes image parts with strong alignment to specific concepts through the most appropriate LoRA experts. This allows each expert to gradually specialize in extracting features for a subset of the \(\mathbf{P}\) concepts, enhancing interpretability and ensuring stronger consistency between classification outcomes and human-understandable visual cues. Although this innovation allows the model to extract a more diverse range of information in few-shot scenarios, it still falls short of being fully optimized. Most deep learning models, when converting from pixel space to semantic space, discard a portion of the pixel-level information to perform efficient computation and recognition. However, in few-shot settings, even the pixel information that is typically discarded is crucial. Therefore, to fully exploit the available information as prompt to activate the pre-trained model’s generalization capabilities, we introduce a Multi-level Feature Aggregation module.

\subsection{\textbf{Multi-level Feature Aggregation}}
\label{sec:multi_feat_agg}
Modern deep learning models often overlook critical fine-grained details in favor of high-level semantic representations, a limitation that is particularly detrimental in few-shot scenarios where each sample must contribute maximally to discriminative features. To alleviate this deficiency in FSPCC, we propose a Multi-level Feature Aggregation module, which integrates representations from different depths of the backbone (low-, mid-, and high-level layers) with the final network output. As illustrated in Fig.\,\ref{fig:ppc_flowchart2}, this module effectively captures complementary information across multiple scales, from local textures to semantic abstractions, thereby improving the feature utilization.

Let \(\mathbf{Z}_L, \mathbf{Z}_M, \mathbf{Z}_H \in \mathbb{R}^{N (\times K) \times R \times D}\) denote feature maps extracted from early (low-level), intermediate (mid-level), and later (high-level) blocks of the backbone, respectively. We also consider the final output feature \(\mathbf{Z}_O \in \mathbb{R}^{N (\times K) \times R \times D}\), which encodes the global semantic representation after the last block. These four feature maps collectively provide a comprehensive view of the input image \(\mathbf{X}\). We begin by performing channel and spatial recalibration on the pair \((\mathbf{Z}_L, \mathbf{Z}_O)\), as defined in Eq.~\ref{eq:ZC}–\ref{eq:EL}, and subsequently apply the same procedure to \((\mathbf{Z}_M, \mathbf{Z}_O)\) and \((\mathbf{Z}_H, \mathbf{Z}_O)\). Specifically, we concatenate \(\mathbf{Z}_L\) and \(\mathbf{Z}_O\) as follows:  
\begin{equation}
\label{eq:ZC}
    \mathbf{Z}_L^{'} = \text{Concat}(\mathbf{Z}_L,\,\mathbf{Z}_O)\,\in \mathbb{R}^{N (\times K) \times R \times 2D}.
\end{equation}

Next, we reshape \(\mathbf{Z}_L^{'}\) to a 4D tensor for spatial operations:
\begin{equation}
    \mathbf{Z}_L^{''} = \text{reshape}(\mathbf{Z}_L^{'})\,\in \mathbb{R}^{N (\times K) \times H \times W \times 2D},
\end{equation}
where \(R = H \times W\) is the total number of patches (or parts) in a 2D arrangement. 

To highlight discriminative structures across channels, we apply global average pooling over spatial positions followed by a \(1\times1\) convolution and sigmoid:
\begin{equation}
\small
    \mathbf{U}_L^c = \mathrm{\text{sigmoid}}\Bigl(\mathrm{conv}_{1 \times 1}\bigl(\mathrm{avgpool}_{\text{spatial}}(\mathbf{Z}_L')\bigr)\Bigr).
\end{equation}

In parallel, we exploit spatial attention by globally averaging features across channels, then applying a \(3\times3\) convolution and sigmoid:
\begin{equation}
\small
    \mathbf{U}_L^s= \mathrm{\text{sigmoid}}\Bigl(\mathrm{conv}_{3 \times 3}\bigl(\mathrm{avgpool}_{\text{channel}}(\mathbf{Z}_L')\bigr)\Bigr).
\end{equation}

By combining these two attention maps multiplicatively, we obtain:
\begin{equation}
    \mathbf{U}_L = \mathbf{U}_L^s \,\times\, \mathbf{U}_L^c \,\in \mathbb{R}^{N (\times K) \times H \times W \times D}.
\end{equation}

Next, we apply \(\mathbf{U}\) to the low-level feature \(\mathbf{Z}_L\) (reshaped in the same way as \(\mathbf{Z}_C'\)) to emphasize informative channels and spatial locations:
\begin{equation}
    \tilde{\mathbf{Z}}_L = \mathbf{U}_L \,\odot\, \mathbf{Z}_L, 
\end{equation}
\begin{equation}
    \mathbf{E}_L = \mathrm{reshape}\bigl(\mathbf{Z}_L + \tilde{\mathbf{Z}}_L\bigr) \in \mathbb{R}^{N (\times K) \times R \times D}.
    \label{eq:EL}
\end{equation}

The result \(\mathbf{E}_L\) is an enriched low-level feature that incorporates global context \(\mathbf{Z}_O\) while preserving valuable fine-grained cues. We repeat the same process for \((\mathbf{Z}_M, \mathbf{Z}_O)\) and \((\mathbf{Z}_H, \mathbf{Z}_O)\) to generate \(\mathbf{E}_M\) and \(\mathbf{E}_H\), which enhance the mid- and high-level features, respectively. Finally, the fused representations are then concatenated to form a unified multilevel embedding:
\begin{equation}
\small
    \mathbf{E}_{All} = \text{Concat}\bigl(\mathbf{E}_L,\,\mathbf{E}_M,\,\mathbf{E}_H\bigr)\,\in \mathbb{R}^{N (\times K) \times R \times 3D}.
\end{equation}

Although \(\mathbf{E}_{All}\) already aggregates features at multiple depths, we further refine it via a lightweight MLP $\textbf{M}$ and Layer Normalization:
\begin{equation}
\small
    \mathbf{E}'_{All} = \mathrm{\text{LN}}\Bigl(\mathrm{\textbf{M}}\bigl(\mathrm{avgpool}(\mathbf{E}_{All})\bigr)\Bigr)\,\in \mathbb{R}^{N (\times K) \times C}.
\end{equation}

Finally, the aggregated feature \(\mathbf{E}'_{All}\) is added to the model’s concept-related feature \(\mathbf{h}\) in Eq.~\ref{h_eq}, capturing both high-level semantics and local details, which is vital for distinguishing classes in data-scarce cenarios.:
\begin{equation}
    \mathbf{F} = \mathbf{h} + \mathbf{E}'_{All} \,\in \mathbb{R}^{N (\times K) \times C}.
\end{equation}

In summary, the Multi-level Feature Aggregation module enriches the feature space of the pre-trained few-shot learner by seamlessly integrating low-, mid-, and high-level features with the final network output. This design significantly enhances generalization to novel classes, as confirmed by our experimental results, where the incorporation of multiscale cues proves highly beneficial for downstream tasks. However, while utilizing representations from multiple depths effectively captures complementary patterns, it also necessitates explicit mechanisms to preserve clear semantic boundaries among the learned concepts. Such mechanisms are critical to maintaining distinct and interpretable regions within the embedding space, thereby safeguarding few-shot performance. To address this challenge, we introduce a Concept Discrimination Loss, which ensures that each concept remains well-separated and interpretable.

\subsection{\textbf{Concept Discrimination Loss}}
\label{sec:cdloss}
To reinforce the separability of concepts in few-shot scenarios, we introduce a Concept Discrimination Loss $\mathcal{L}_{CD}$ that drives each learned concept to occupy a distinct region of the semantic manifold. Unlike traditional classification-centric constraints, this loss explicitly encourages minimal overlap among different concepts by imposing a strong penalty when concept activations become overly entangled. As a result, the features associated with each concept remain both semantically cohesive and mutually exclusive, thereby facilitating transparent concept-level interpretations.

Let \(\mathbf{A} = [A_{1}, A_{2}, \dots, A_{C}]\) represent the concept activation vectors generated by the PCL in Section~\ref{sec:FSPCC}. To enforce separability among these concepts, we employ the InfoNCE loss\cite{oord2018representation} with temperature parameter \(\kappa\) to derive a probability distribution over the \(C\) concepts. The $\mathcal{L}_{\text{CD}}$ is defined as:
\begin{equation}
    \mathcal{L}_{\text{CD}} = -\frac{1}{|C|} \sum_{i \in C} \log \frac{\exp(A_i/ \kappa)}{\sum_{k=1}^{C} \exp(A_k/ \kappa)},
\label{eq:cdloss}
\end{equation}
where \(\kappa\) governs the sharpness of the distribution. A smaller \(\kappa\) forces the network to assign high activation to a dominant concept while suppressing others, enhancing concept discriminability. Conversely, a larger \(\kappa\) yields smoother distributions, allowing moderate concept overlap to capture richer semantics.

\subsection{\textbf{Training and Testing}}
During the training phase, we feed each image into the backbone enhanced with multi-level feature aggregation module. This process yields enriched representations that are further projected into concept space through the PCL module. Two principal objectives then guide network optimization. First, we enforce the concept discrimination loss \(\mathcal{L}_{CD}\) in Section~\ref{sec:cdloss} to encourage separability among the learned concepts, thereby minimizing potential overlaps within the concepts. Second, to accommodate the few-shot classification task, we employ an episodic paradigm with support and query sets, following \(N\)-way \(K\)-shot protocols. Specifically, after computing class-specific prototypical concepts from the support samples, each query sample’s label is predicted by \(\mathbf{Y}_q = \arg \max (\mathbf{M})\), as stated in Section~\ref{sec:FSPCC}. We then compute a standard cross-entropy loss \(\mathcal{L}_{CLS}\) based on the predicted labels and ground truths for all query samples. Consequently, the network learns not only to classify in a human-interpretable manner under few-shot conditions but also to preserve distinctive concepts.

At inference time, we freeze all model parameters and perform few-shot classification on novel classes that are unseen during training. Given a small number of labeled samples per novel class (i.e., the support set), we derive their prototypical concepts in the learned concept space. A query image is then processed through the same feature extractor to obtain its concept activations, which are subsequently compared to the class-level prototypical support concepts. The classification decision is based on the highest cosine similarity value across the candidate classes. 


\begin{table}[!ht]
    \centering
    \caption{Summary of FSL benchmark datasets.}
    \label{tab:dataset_summary}
    \resizebox{\columnwidth}{!}{%
    \begin{tabular}{c c c c c}
    \toprule
    \textbf{Dataset} & \textbf{Domain} & \textbf{Classes} & \textbf{Train} & \textbf{Test} \\ 
    \midrule
    CUB-200-2011 & Birds & 200 & 9374 & 2404 \\ 
    \emph{mini}-ImageNet & Natural Images & 100 & 38400 & 12000 \\ 
    CIFAR-FS & Objects & 100 & 38400 & 12,000 \\ 
    Stanford Cars & Vehicles & 196 & 10722 & 4021 \\ 
    FGVC-Aircraft & Aircraft & 100 & 8000 & 2000 \\ 
    DTD & Textures & 47 & 3,960 & 1,680 \\ 
    \bottomrule
    \end{tabular}
    }
\end{table}

\section{Experiments}
\label{sec4}
\textbf{Datasets.} To holistically evaluate our model generalization in FSPCC, we systematically curate six benchmark datasets spanning distinct visual domains with controlled heterogeneity. The evaluation suite encompasses: 1) Fine-grained recognition through \textit{CUB-200-2011}~\cite{WahCUB_200_2011} (200 avian species, 11,788 images) featuring high inter-class similarity with part-level annotations, and \textit{FGVC-Aircraft}~\cite{maji2013fine} (100 aircraft models) capturing manufacturer-induced structural variations; 2) Generic image classification via \textit{mini-ImageNet}~\cite{vinyals2016matching} (100 classes, 600 samples/class) and \textit{CIFAR-FS}~\cite{bertinetto2018meta} (100 classes), providing standardized benchmarks for cross-domain few-shot scenarios; 3) Specialized domain adaptation using \textit{Stanford Cars}~\cite{hilliard2018few} (196 vehicle categories) with manufacturer-specific intra-class variances and \textit{DTD}~\cite{cimpoi2014describing} (47 texture classes) modeling human-perceived material attributes. This intentionally heterogeneous collection introduces orthogonal challenges in semantic granularity (47–200 classes), resolution regimes (32×32 to 224×224 pixels), and domain shifts, rigorously probing model robustness. More detailed information is provided in Tab.~\ref{tab:dataset_summary}.

\textbf{Experiment Setup} 
In our experimental setup, we adopt the DINO as the backbone. The parameter-efficient LoRA experts in PCM are configured with a unified rank of 8, maintaining an $\alpha$ scaling factor of 32 alongside a dropout probability of 0.1 to ensure robust feature representation. The optimization process employs the AdamW algorithm with a base learning rate of 1e-2, incorporating a multi-stage learning policy comprising 15 warm-up epochs for gradual parameter adaptation followed by cosine annealing for convergence refinement. The complete training regimen spans 80 epochs with 500 episodic iterations per epoch, ensuring comprehensive task exposure and model stabilization. All implementations and evaluations are executed on a singular NVIDIA A6000 GPU accelerator, adhering to consistent hardware specifications for controlled experimental comparisons.

\begin{table*}[!ht]
    \centering
    \caption{Classification accuracy (\%) of FSL methods evaluated on the CUB, Stanford Cars, and Aircraft datasets across 1-shot, 5-shot, 10-shot, and 20-shot tasks.}
    \label{table2}
    \arrayrulecolor{black}
    \definecolor{lightpurple}{RGB}{240, 230, 255} 

    \resizebox{\textwidth}{!}{
    \begin{tabular}{c|cccc|cccc|cccc}
        \toprule
        Method & \multicolumn{4}{c|}{CUB} & \multicolumn{4}{c|}{Stanford Cars} & \multicolumn{4}{c}{Aircraft} \\ 
        ~ & 1-shot & 5-shot & 10-shot & 20-shot & 1-shot & 5-shot & 10-shot & 20-shot & 1-shot & 5-shot & 10-shot & 20-shot \\ 
        \midrule
        ProtoPNet\cite{chen2019looks} & 63.57 & 74.20 & 75.31 & 76.29 & 60.39 & 77.10 & 77.76 & 78.95 & 59.16 & 69.87 & 70.46 & 71.42 \\ 
        TesNet\cite{wang2021interpretable} & 65.32 & 74.01 & 76.27 & 77.15 & 63.67 & 71.49 & 72.65 & 73.38 & 62.33 & 70.82 & 71.68 & 72.39 \\
        Eval-ProtoPNet\cite{huang2023evaluation} & 64.21 & 73.29 & 75.43 & 77.68 & 57.54 & 81.95 & 84.83 & 85.46 & 55.85 & 79.41 & 80.83 & 82.85 \\ 
        ST-ProtoPNet\cite{wang2023learning} & 56.19 & 63.73 & 66.33 & 67.87 & 73.94 & \underline{88.65} & \underline{90.35} & \underline{91.02} & \underline{78.57} & \underline{87.21} & \underline{89.48} & \underline{90.04} \\ 
        ProtosViT\cite{turbe2024protos} & \underline{71.81} & \underline{86.12} & \underline{89.83} & \underline{90.47} & \underline{82.95} & 86.57 & 86.94 & 88.06 & 77.67 & 86.91 & 88.72 & 89.66 \\ 
        \rowcolor{lightpurple}
        LE-ProtoPNet (Ours) & \textbf{83.43} & \textbf{92.07} & \textbf{93.42} & \textbf{94.16} & \textbf{85.38} & \textbf{92.20} & \textbf{93.87} & \textbf{94.41} & \textbf{82.82} & \textbf{89.13} & \textbf{90.35} & \textbf{91.60} \\ 
        \bottomrule
    \end{tabular}
    }
\end{table*}

\begin{table*}[!ht]
    \centering
    \caption{Classification accuracy (\%) of FSL methods evaluated on the \emph{mini}-ImageNet, CIFAR-FS, and DTD datasets across 1-shot, 5-shot, 10-shot, and 20-shot tasks.}
    \label{table3}
    \arrayrulecolor{black}
    \definecolor{lightpurple}{RGB}{240, 230, 255} 
    \resizebox{\textwidth}{!}{
    \begin{tabular}{c|cccc|cccc|cccc}
        \toprule
        Method & \multicolumn{4}{c|}{\emph{mini}-ImageNet} & \multicolumn{4}{c|}{CIFAR-FS} & \multicolumn{4}{c}{DTD} \\
        ~ & 1-shot & 5-shot & 10-shot & 20-shot & 1-shot & 5-shot & 10-shot & 20-shot & 1-shot & 5-shot & 10-shot & 20-shot \\ 
        \midrule
        ProtoPNet\cite{chen2019looks} & 50.28 & 67.27 & 69.39 & 70.41 & 52.86 & 71.43 & 72.62 & 73.45 & 45.69 & 57.97 & 58.56& 60.21 \\ 
        TesNet\cite{wang2021interpretable} & 58.58 & 83.66 & 86.25 & 86.40 & 55.85 & 72.29 & 73.71 & 73.76 & 44.83 & 54.55 & 56.03 & 57.87 \\ 
        Eval-ProtoPNet\cite{huang2023evaluation} & 60.50 & \textbf{87.38} & \textbf{89.98} & \textbf{90.78} & 44.64 & 65.40 & 68.02 & 69.54 & 54.43 & 73.03 & 74.99 & 76.50 \\    
        ST-ProtoPNet\cite{wang2023learning} & \underline{70.53} & 83.30 & 85.97 & 86.45 & 62.44 & 77.07 & 79.76 & 81.21 & \underline{55.29} & \underline{73.20} & \underline{75.30} & \underline{77.26} \\ 
        ProtosViT\cite{turbe2024protos} & 68.82 & 82.78 & 85.36 & 86.45 & \underline{76.65} & \underline{79.67} & \underline{81.20} & \underline{82.37} & 51.21 & 67.53 & 70.72 & 72.89\\ 
        \rowcolor{lightpurple}
        LE-ProtoPNet (Ours) & \textbf{81.34} & \underline{87.17} & \underline{89.66} & \underline{90.13} & \textbf{79.69} & \textbf{86.40} & \textbf{87.50} & \textbf{88.47} & \textbf{60.05} & \textbf{76.15} & \textbf{78.41} & \textbf{80.22} \\ 
        \bottomrule
    \end{tabular}
    }
\end{table*}

\subsection{General Few-Shot Classification}
The performance of our LE-ProtoPNet, as shown in Tab.~\ref{table2} and Tab.~\ref{table3}, demonstrates its advantage over the comparison methods, with notable improvements in accuracy across multiple datasets, including CUB, Stanford Cars, Aircraft, \emph{mini}-ImageNet, CIFAR-FS, and DTD. We use bold to indicate the optimal results and underline to represent the sub-optimal results. These tables highlight our LE-ProtoPNet’s ability to handle both fine-grained and coarse-grained datasets, showcasing its robust generalization capabilities.

In the fine-grained datasets, as detailed in Tab.~\ref{table2}, our LE-ProtoPNet consistently outperforms the comparison methods, including ProtoPNet\cite{chen2019looks}, TesNet\cite{wang2021interpretable}, Eval-ProtoPNet\cite{huang2023evaluation}, ST-ProtoPNet\cite{wang2023learning} and ProtosViT\cite{turbe2024protos} on the CUB, Stanford Cars and Aircraft datasets. Specifically, on CUB, the LE-ProtoPNet achieves 83.43\% in 1-shot and 92.07\% in 5-shot, establishing a significant lead over other methods. Similarly, on Stanford Cars and Aircraft, our LE-ProtoPNet outperforms most competitors, demonstrating its robustness in handling subtle class distinctions. This strong performance underscores the ability of the LE-ProtoPNet to effectively capture detailed feature representations and align them with concepts, which is crucial for distinguishing between closely related categories in fine-grained classification tasks.

On coarse-grained datasets, as shown in Tab.~\ref{table3}, our LE-ProtoPNet also performs exceptionally well, achieving 1-shot accuracy of 81.34\% on \emph{mini}-ImageNet and 79.69\% on CIFAR-FS, and 5-shot accuracy of 87.17\% and 86.40\%, respectively. These results demonstrate that our LE-ProtoPNet is highly effective even in less granular tasks, where the challenge lies in distinguishing broader categories. However, while our LE-ProtoPNet performs competitively across these datasets, it still slightly trails behind Eval-ProtoPNet\cite{huang2023evaluation} in certain settings. This performance difference can be attributed to the design of our approach, which is optimized for capturing detailed and discriminative features, especially in fine-grained tasks. In coarse-grained datasets like \emph{mini}-ImageNet, where the visual distinctions between categories are less subtle, our LE-ProtoPNet may not fully exploit the simpler features required for effective classification. As a result, while our LE-ProtoPNet excels in fine-grained scenarios, there is room for further refinement in coarser tasks, where models such as Eval-ProtoPNet\cite{huang2023evaluation} seem to better capture broader feature representations.

For the DTD dataset,
the proposed LE-ProtoPNet outperforms all other comparison methods. This strong performance indicates that our approach, despite its primary optimization for object-centric tasks, is capable of effectively handling texture-based recognition as well. The superior accuracy can be attributed to the generalization of our LE-ProtoPNet, which allows the model to engage the most relevant features for texture differentiation. While DTD presents challenges due to the complex nature of texture patterns, our LE-ProtoPNet's ability to capture these nuances showcases its robustness and versatility.


\begin{table*}[!ht]
    \centering
    \caption{Effectiveness of our proposed modules on classification accuracy (\%) on CUB Dataset: Ablation Study.}
    \label{table4}
    \arrayrulecolor{black}
    \definecolor{lightpurple}{RGB}{240, 230, 255} 
    \begin{tabular}{ccccccc|cc}
    \toprule
        Linear Probe&FFT&LoRA&MoLE& PCM & MFA & \(\mathcal{L}_{\mathrm{CD}}\) & 1-shot &5-shot  \\ 
    \midrule
    \checkmark&~ &~ & ~ & ~ & ~ & ~ & 70.44 & 78.18 \\ 
    ~ & \checkmark & ~ &~ & ~ & ~ & ~ & 71.81 & 86.12 \\
       ~ & ~ & \checkmark & ~ & ~ & ~ & ~ & 77.96 & 87.74 \\ 
       ~ & ~ & ~ & \checkmark & ~ & ~ & ~ & 78.53 & 90.15 \\ 
       ~ & ~ & ~ & \checkmark& \checkmark & ~ & ~ & 80.96 & 91.10  \\ 
       ~ & ~ & ~ & \checkmark& \checkmark & \checkmark & ~ & 81.62 & 91.80 \\ 
        ~ & ~ & ~ & \checkmark & \checkmark & ~ &\checkmark & 79.80 & 91.73 \\ 
       ~ & ~ & ~ & \checkmark& ~ & \checkmark &\checkmark & 82.21 &  91.40 \\ 
       \rowcolor{lightpurple}
       ~ & ~  & ~ & \checkmark & \checkmark & \checkmark & \checkmark & \textbf{83.43} & \textbf{92.07} \\  
    \bottomrule
    \end{tabular}
\end{table*}

\subsection{Ablation Study} In this section, we discuss in detail the effectiveness of our proposed modules. We complete the ablation study on the CUB dataset with 5-way 1-shot and 5-way 5-shot tasks. Experimental results are presented in Tab.~\ref{table4}, where \checkmark represents presence of the module.

\subsubsection{Impact of Prototypical Concept-guided MoLE} The PCM enhances few-shot classification by activating the strong generalization capabilities of pre-trained models using experts guided by concepts. This subsection investigates the impact of the PCM on model performance by comparing it with other fine-tuning methods. The introduction of PCM significantly improves performance compared to linear probe, fully fine-tune (FFT), standard LoRA and MoLE. While MoLE achieves notable gains over LoRA by integrating mixture-of-experts tuning, the lack of concept guidance limits its capability to focus on part-specific knowledge. By incorporating PCL, PCM leverages concept-representative information, leading to improved accuracy (e.g. 80. 96\% vs 78.53\% in the 1-shot setting and 91.10\% vs 90.15\% in the 5-shot setting). This highlights the critical role of concept guidance in aligning the model's representations with the underlying data distribution.

\subsubsection{Impact of Multilevel Feature Aggregation (MFA)} MFA aggregates features from multiple layers, preserving both low-level details and high-level semantic features. This approach ensures that the model retains more comprehensive and discriminative information, which is particularly important in FSL. As demonstrated in our experiments, the addition of MFA leads to substantial improvements in both 1-shot and 5-shot performance. Specifically, when MFA is integrated with PCM, the model achieves higher accuracy compared to PCM alone, with significant improvements in classification performance. This demonstrates that MFA effectively enhances feature representation, especially in challenging few-shot scenarios, by preventing the loss of critical information from the final layers.

\subsubsection{Impact of Concept Discrimination Loss}As shown in our experimental results, when combined with PCM and MFA, \(\mathcal{L}_{CD}\) ensures that prototypes maintain clear boundaries, reducing the risk of confusion during inference. This results in higher accuracy in both 1-shot and 5-shot settings, with the best overall performance achieved when all three components PCM, MFA, and \(\mathcal{L}_{CD}\) are used together. This demonstrates that \(\mathcal{L}_{CD}\) is crucial for improving the robustness of the model, leading to more accurate and reliable predictions in few-shot learning.


\subsection{Hyperparameter Analysis}
In this section, we conduct a thorough hyperparameter analysis to assess the impact of key factors on model performance. Specifically, we investigate the influence of the number of experts \(\mathbf{E}\), the number of prototypes \(\mathbf{C}\), and the coefficient \(\lambda\) of \(\mathcal{L}_{CD}\). By systematically varying these parameters, we aim to identify their optimal configurations and understand how each factor contributes to the model's ability to effectively learn and generalize across different tasks. 

\begin{table}[!ht]
    \centering
    \caption{Study on the number of concepts on \emph{mini}-ImageNet (\%).}
    \label{table5}
    \arrayrulecolor{black}
    \begin{tabular}{c|cccc}
    \toprule
        Nubmer of Concepts & 1-shot&5-shot& 10shot & 20shot\\ 
    \midrule
       100& 73.00 & 84.74 & 86.73& 87.69 \\
       312& \textbf{81.34} & \textbf{87.17} & \textbf{89.66} & \textbf{90.13}\\
       400& 72.85 & 86.29 & 89.11 &89.65 \\
       500& 72.93 & 84.63 & 87.27 & 88.19 \\
       1000& 74.51 & 86.91 & 89.12 & 89.86 \\ 
    \bottomrule
    \end{tabular}
\end{table}


\subsubsection{Number of Concepts}Based on the experimental results in Tab.~\ref{table5}, the performance of the model on the \emph{mini}-ImageNet dataset demonstrates that 312 concepts yield the best results in both 1-shot, 5-shot, 10-shot and 20-shot settings. Specifically, in the 1-shot scenario, the accuracy peaks at 81.34\% with 312 concepts, and in the 5-shot setting, it reaches 87.17\%. As the number of concepts increases to 400, 500, and 1000, the accuracy either decreases or remains below the performance achieved with 312 concepts. Notably, for 1000 concepts, the accuracy in the 1-shot and 5-shot settings is lower compared to 312 concepts, which suggests that further increasing the number of concepts does not provide additional benefits. These findings suggest that 312 concepts offer an optimal balance between the richness of the learned features and the model’s ability to generalize. Therefore, we configure 312 concepts as our experimental configuration to ensure optimal performance. 

\begin{table}[!ht]
    \centering
    \caption{Study on the number of experts on \emph{mini}-ImageNet (\%).}
    \label{table6}
    \arrayrulecolor{black}
    \begin{tabular}{c|cccc}
    \toprule
        Nubmer of Experts & 1-shot&5-shot&10-shot&20shot\\ 
    \midrule
       1& 73.49 & 85.65 & 87.53 & 89.03 \\
       3& \textbf{81.34} & \textbf{87.17}& \textbf{89.66} &\textbf{90.13}\\
       6& 72.94 & 82.92 & 85.96 &87.05\\
       8 & 73.56 & 83.11& 86.08 & 86.98 \\
      
    \bottomrule
    \end{tabular}
\end{table}

\subsubsection{Number of Experts}
In this experiment, we investigate the effect of varying the number of experts \(\mathbf{E}\) on the performance of our model. The number of experts plays a crucial role in determining the model's capacity for learning diverse and specialized features. By adjusting \(\mathbf{E}\), we can control the model's flexibility to allocate different experts to handle various aspects of the data. Thus, we select 4 values of \(\mathbf{E}\) in our LE-ProtoPNet for analysis. The experimental results, presented in Tab.~\ref{table6}, reveal that setting \(\mathbf{E}\) to 3 yields the best performance across all shot settings.In this situation, the model strikes an optimal balance, leveraging enough experts to capture diverse feature representations without overcomplicating the model. This enables it to generalize effectively across different tasks. When \(\mathbf{E}\) is set to 1, the model behaves similarly to a standard LoRA, resulting in relatively lower performance. Furthermore, when \(\mathbf{E}\) exceeds 3, the accuracy begins to decline, indicating that larger values of \(\mathbf{E}\) may lead to diminishing returns. 

\begin{figure}[h]
    \begin{flushright}
        \centering
        \includegraphics[scale=0.3]{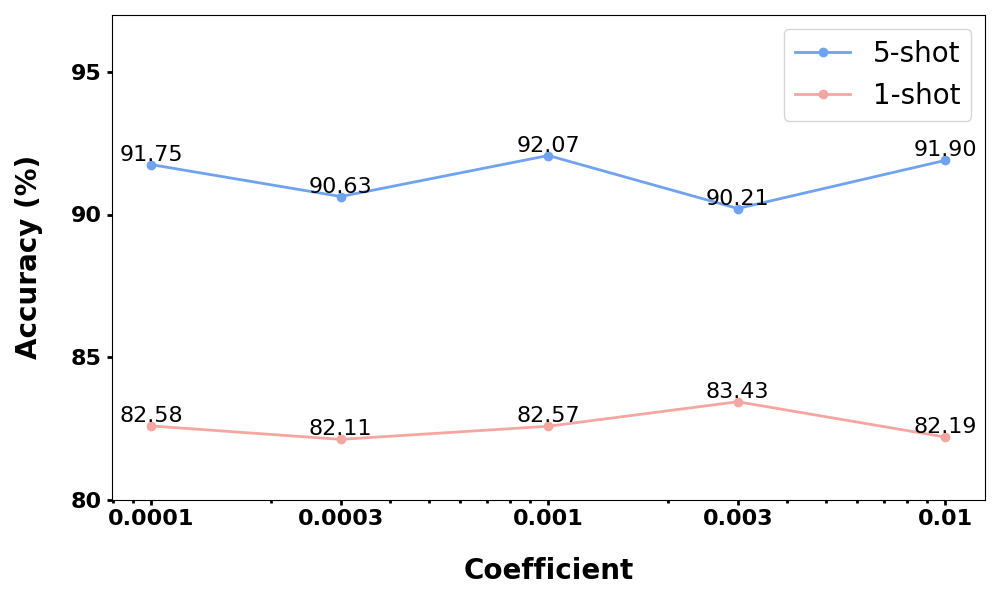} 
        \centering
        \caption{Study on the impact of coefficient \(\lambda\) of \(\mathcal{L}_{CD}\) on CUB.}
        \label{fig5}
    \end{flushright}
\end{figure}

\subsubsection{Coefficient \(\lambda\) of \(\mathcal{L}_{CD}\)}
We investigated the impact of the coefficient \(\lambda\) of \(\mathcal{L}_{CD}\) on the model's performance through experiments, with the specific results presented in Fig.~\ref{fig5}. \(\lambda\)  is essential for balancing the trade-off between the discrimination and the representational ability of the prototypical concepts. We observe that the model’s accuracy  fluctuates as the coefficient \(\lambda\) changes. Specifically, performance peaks at \(\lambda\)= 0.001 in 5-shot setting where accuracy reaches 92.07\% and \(\lambda\)= 0.003 in 1-shot setting where accuracy achieves 83.43\%. Increasing \(\lambda\) beyond these optimal points leads to a decline in accuracy, with over-regularization likely hindering the model’s generalization ability, particularly in the 1-shot setting.  


\subsection{Interpretability}
In this section, we explore the interpretability of our model. Standard ProtoPNet\cite{chen2019looks} enables users to visualize the concepts and their corresponding locations on the images. Additionally, it facilitates the identification of the training images that are most similar to the concepts, thereby providing an explanation for the classification decisions through a ``this looks like that'' approach. For instance, if the concept captures a specific part of a bird, such as its wings, users can visualize the concept along with the training images that most closely match it, thereby gaining insight into how this particular feature influences the model's classification decision. Our LE-ProtoPNet follows a similar explanatory approach, but with a key difference: we visualize the activations of the concepts on both the query images and the corresponding support images simultaneously. This provides a more comprehensive understanding of the model’s decision-making process for few-shot classification by comparing the similarities and differences of the activation maps of the concepts across images in both sets. If the concepts shared between query image and support image are more numerous and the similarity of the activation maps in the corresponding regions is higher, we can infer that the model is likely to classify them as belonging to the same category.

\begin{figure*}
  \centering
  \includegraphics[scale=0.8]{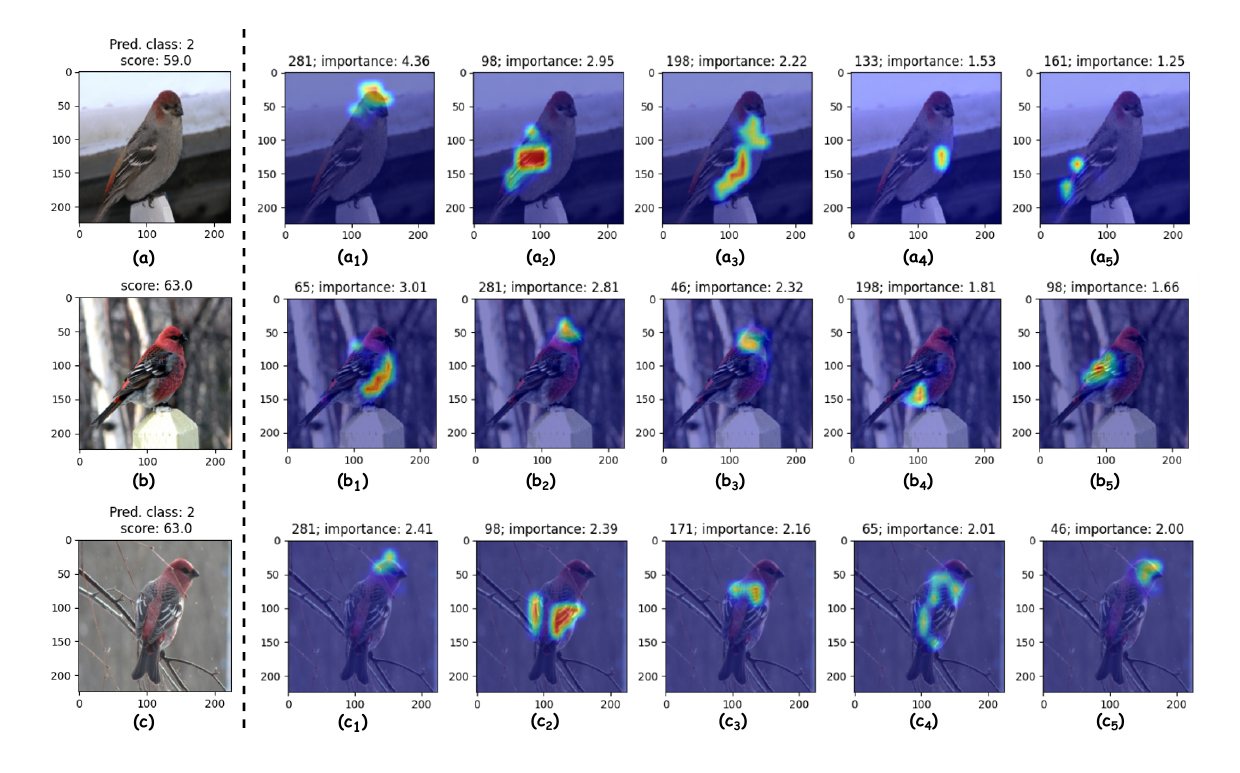} 
  \centering
  \caption{An example of the interpretability of our model demonstrated on CUB dataset including a support image and two query images.}
  \label{fig4}
\end{figure*}

Figure~\ref{fig4} illustrates an example of the interpretability of our LE-ProtoPNet. Here, \(a\) \(c\) refer to the images from the query set, while \(b\) refers to the image from the support set. It can be seen that the top-5 most important concepts in the support set are numbered 65, 281, 46, 198, and 98, whose visual semantics can roughly be interpreted as `red feather', `red head', `black frame around the eyes', `white flank', and `Black and white wings', respectively. This indicates that the PCL module is capable of learning useful and interpretable feature components, even with a limited number of samples. The label ``Pred. class: 2'' above the images \(a\) \(c\) indicates that both images \(a\) \(c\) from the query set belong to the same class as the image \(b\) from the support set , as \(b\) is assigned the number 2 in the support set. The ``score'' represents the similarity score between the two \(a\) \(c\) images and \(b\). Our interpretability also follows the ``this looks like that'' approach. Image \(c\) is classified into the same category as image \(b\) because \(c_1\) looks like \(b_2\), \(c_2\) looks like \(b_5\), \(c_4\) looks like \(b_1\), and \(c_5\) looks like \(b_3\). This allows us to understand the rationale behind the model's classification decision. In image \(a\), the concepts 281, 198, and 98 look like those in \(b\). Based on these factors, \(a\) is classified in the same category as \(b\). 
However, due to the viewing angle, the red feathers are nearly invisible in image \(a\), causing concept 65 to not appear. Instead, concept 161 shows a small portion of the red feathers on the tail, and concept 133 displays the white chest, which is barely visible in \(b\). Since the differences between the concepts are greater than in the case of \(c\), the similarity score between \(a\) and \(b\) is lower than that between \(b\) and \(c\), which aligns well with our human intuitive understanding of object recognition.

\section{conclusion}
\label{sec5}
In this paper, we have proposed a novel setting named Few-Shot Prototypical Concept Classification (FSPCC). To overcome parametric imbalance and representation misalignment in FSPCC, we propose our LE-ProtoPNet including four integrated enhancements. Our LE-ProtoPNet outperforms the comparison metheds across a wide range of datasets, demonstrating superior performance on both fine-grained and coarse-grained tasks. Our ablation study demonstrates that each proposed module contributes to the overall improvement in performance. Furthermore, we highlight the interpretability of our model, demonstrating how it aligns visual features with concepts to support its classification decisions. This interpretability allows us to understand the reasoning behind the model’s predictions, making the decision-making process transparent and aligning it with human intuitive recognition.

\bibliographystyle{IEEEtran}
\bibliography{refs}

\vspace{-2ex}
\begin{IEEEbiography}[{\includegraphics[width=1in,height=1.25in,clip,keepaspectratio]{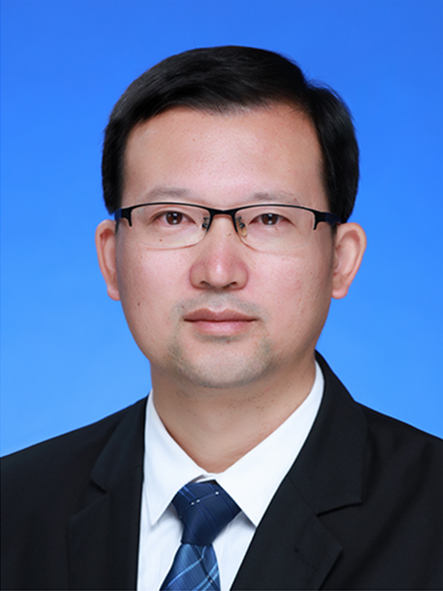}}]{Zhong Ji} received the Ph.D. degree in signal and information processing from Tianjin University, Tianjin, China, in 2008. He is currently a Professor with the School of Electrical and Information Engineering, Tianjin University. He has authored over 100 technical articles in refereed journals and proceedings. His current research interests include continual learning, few shot leanring, and cross-modal analysis.
\end{IEEEbiography}

\vspace{-2ex}
\begin{IEEEbiography}[{\includegraphics[width=1in,height=1.25in,clip,keepaspectratio]{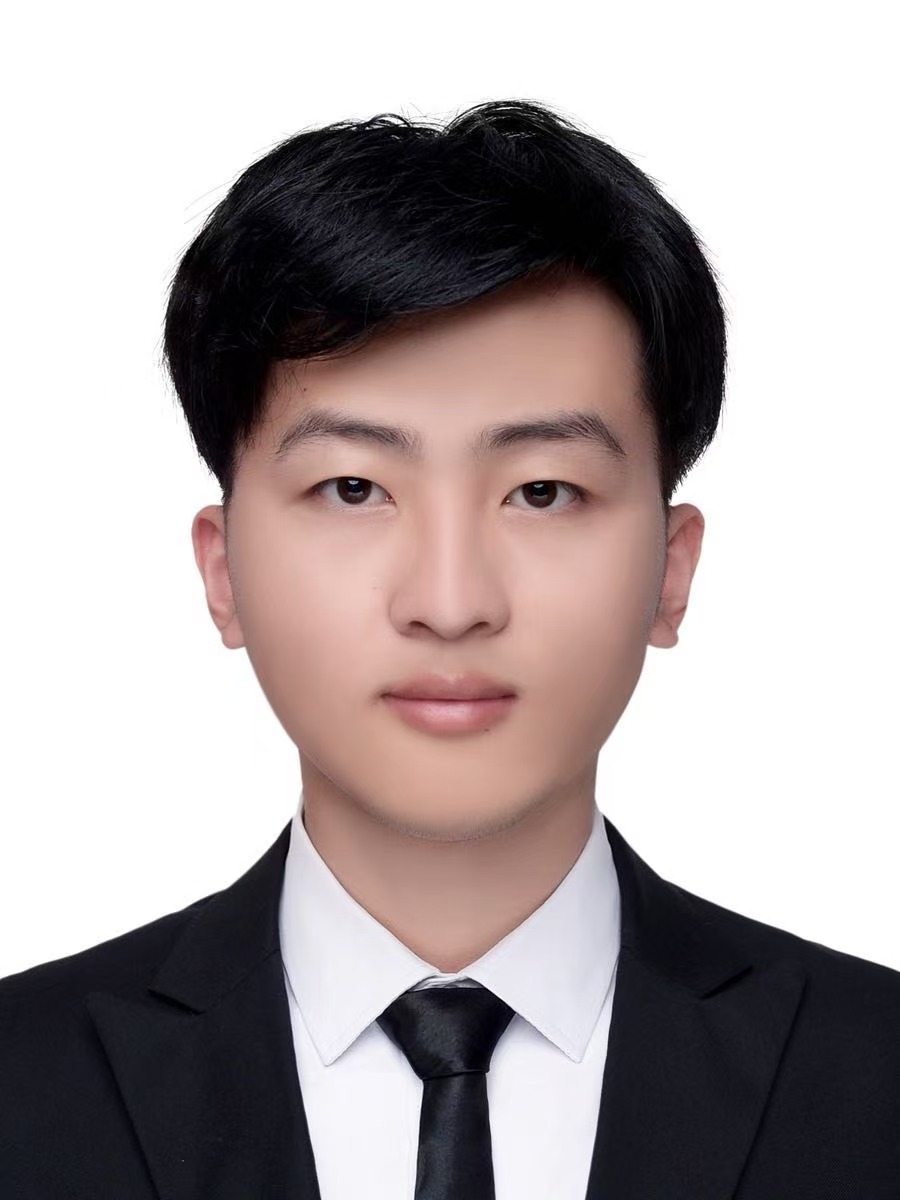}}]{Rongshuai Wei} received the B.S. degree in College of Electronic Science and Engineering, JiLin University, Changchun, China, in 2023, and is currently studying for an M.S. degree in the School of Electrical and Information Engineering, Tianjin University, Tianjin, China. His research focuses on few-shot learning and explainable artificial intelligence.
\end{IEEEbiography}

\vspace{-2ex}
\begin{IEEEbiography}[{\includegraphics[width=1in,height=1.25in,clip,keepaspectratio]{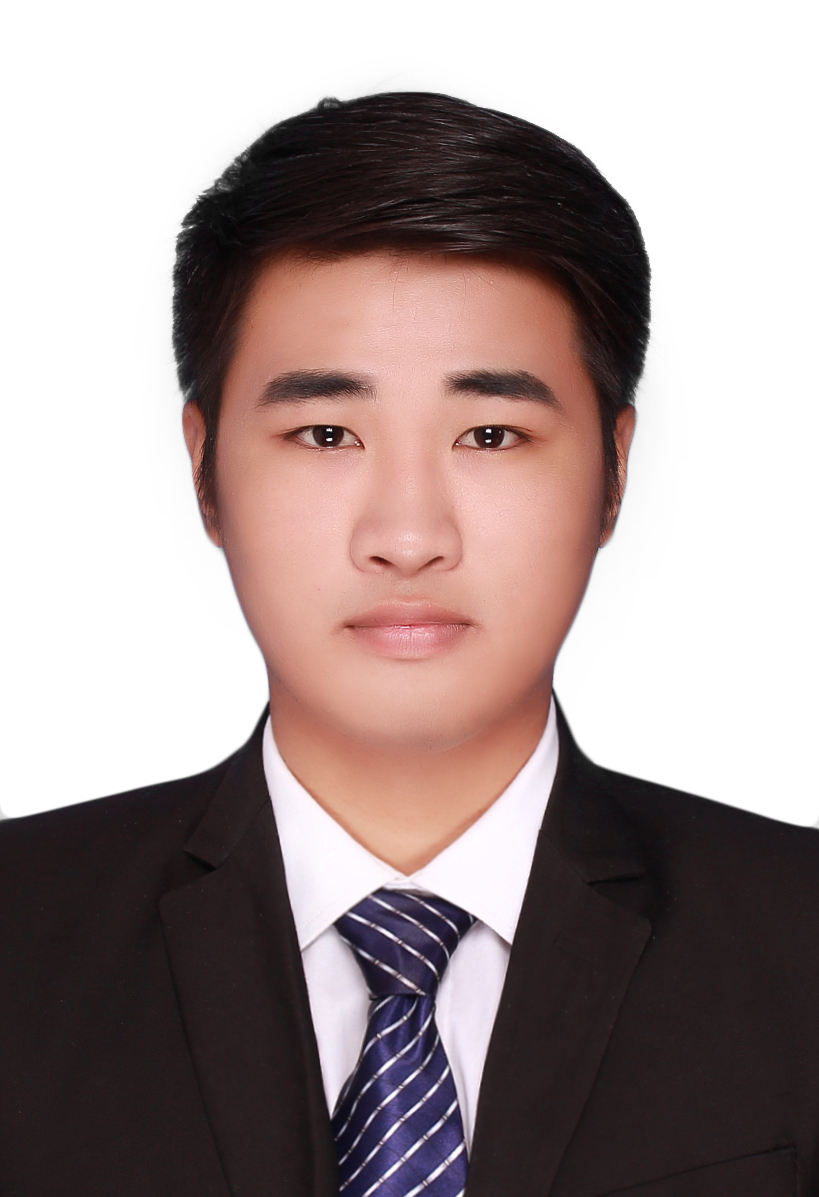}}]{Jingren Liu} received the B.S. degree in Computer Science and Technology from Nanjing University of Finance and Economy, Nanjing, China, in 2019, and is currently working toward the PhD degree in the School of Electrical and Information Engineering, Tianjin University, Tianjin, China. His current research interests include continual learning, few shot leanring, and prompt learning.
\end{IEEEbiography}

\vspace{-2ex}
\begin{IEEEbiography}[{\includegraphics[width=1in,height=1.25in,clip,keepaspectratio]{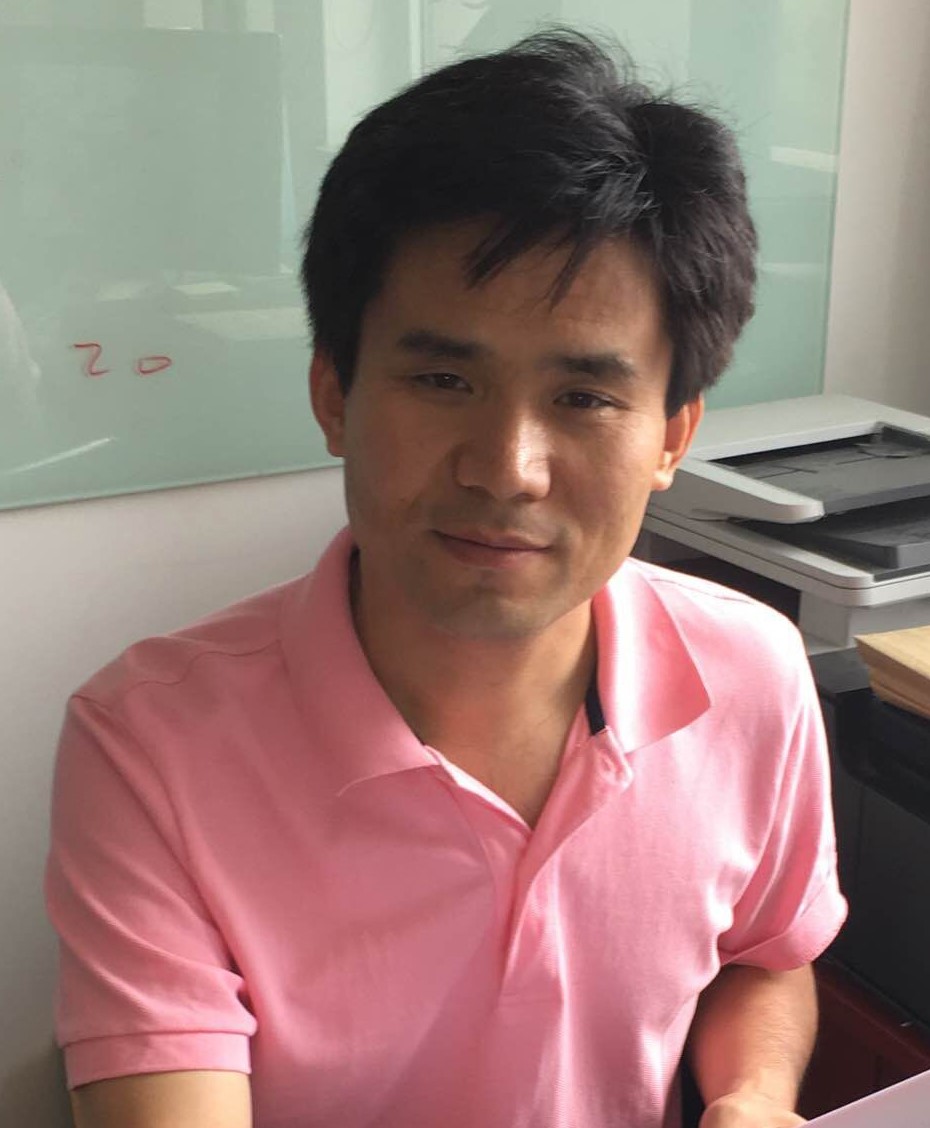}}]{YanWei Pang} received the Ph.D. degree in electronic engineering from the University of Science and Technology of China, Hefei, China, in 2004. He is currently a Professor with the School of Electrical and Information Engineering, Tianjin University, Tianjin, China. He has authored over 200 scientific papers. His current research interests include object detection and recognition, vision in bad weather, and computer vision.
\end{IEEEbiography}

\vspace{-2ex}
\begin{IEEEbiography}[{\includegraphics[width=1in,height=1.25in,clip,keepaspectratio]{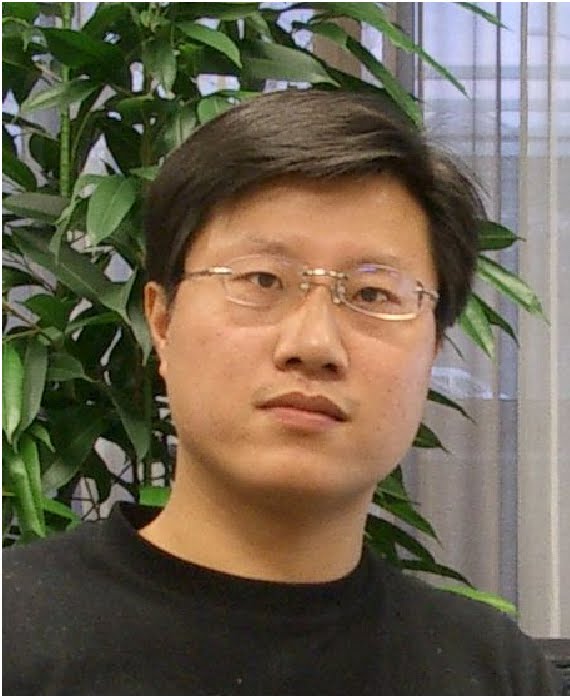}}]{Jungong Han} received the Ph.D. degree in telecommunication and information system from Xidian University, Xi’an, China, in 2004. He is a Professor of the Department of Automation, Tsinghua University, Beijing 100084, China. He has published over 200 articles, including more than 80 IEEE Transactions and more than 50 A* conference articles. His research interests span the fields of video analysis, computer vision, and applied machine learning. He is a Fellow of the International Association of Pattern Recognition.
\end{IEEEbiography}

\vfill
\end{document}